\documentclass{article}
\usepackage{spconf,amsmath,graphicx}
\usepackage{hyperref}
\usepackage{xcolor}
\usepackage{graphicx}      
\usepackage{subcaption}    
\usepackage{multirow}
\usepackage{siunitx}
\usepackage{enumerate}
\usepackage{amssymb}  
\usepackage{pifont}   
\usepackage{etoolbox}
\let\oldthebibliography\thebibliography
\renewcommand{\thebibliography}[1]{
  \oldthebibliography{#1}
  \footnotesize 
  \setlength{\itemsep}{0pt plus 0.3ex}
}

\title{Improving Novel view synthesis of 360$^\circ$ Scenes in Extremely Sparse Views by Jointly Training Hemisphere Sampled Synthetic Images}
\name{Guangan Chen$^1$, Anh Minh Truong$^1$, Hanhe Lin$^2$, Michiel Vlaminck$^1$, Wilfried Philips$^1$, Hiep Luong$^1$}
\address{$^1$Image Processing and Interpretation (IPI), IMEC research group at Ghent University, Belgium\\$^2$School of Science and Engineering, University of Dundee, United Kingdom}

\begin{document}
%
\maketitle
\begin{abstract}
Novel view synthesis in 360$^\circ$ scenes from extremely sparse input views is essential for applications like virtual reality and augmented reality. 
This paper presents a novel framework for novel view synthesis in extremely sparse-view cases. As typical structure-from-motion methods are unable to estimate camera poses in extremely sparse-view cases, we apply DUSt3R to estimate camera poses and generate a dense point cloud. 
Using the poses of estimated cameras, we densely sample additional views from the upper hemisphere space of the scenes, from which we render synthetic images together with the point cloud.
Training 3D Gaussian Splatting model on a combination of reference images from sparse views and densely sampled synthetic images allows a larger scene coverage in 3D space, addressing the overfitting challenge due to the limited input in sparse-view cases. 
Retraining a diffusion-based image enhancement model on our created dataset, we further improve the quality of the point-cloud-rendered images by removing artifacts.
We compare our framework with benchmark methods in cases of only four input views, demonstrating significant improvement in novel view synthesis under extremely sparse-view conditions for 360$^\circ$ scenes. The source code is available at \url{https://github.com/angchen-dev/hemiSparseGS}.
\end{abstract}

\begin{keywords}
3D Gaussian Splatting, 360$^\circ$ scenes, \mbox{extremely sparse views}, diffusion model, image enhancement
\end{keywords}

\begin{figure*}[htp]
    \centering
    \includegraphics[width=0.95\textwidth]{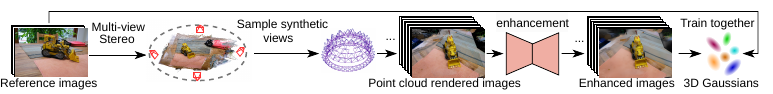}
  \caption{\textbf{The framework of proposed method.} Given a sparse set of views (reference images) with unknown camera poses of a scene, we first utilize an off-the-shelf multi-view stereo method to estimate the camera poses and generate a dense point cloud. Subsequently, we sample camera poses in the upper hemisphere space of the scene and render synthetic images using the sampled poses and the dense point cloud. Next, the rendered images are enhanced using a diffusion-based model. Finally, both the enhanced synthetic images and reference images are employed to train a 3DGS model.}
  \label{fig:framework}
\vspace{-10pt}
\end{figure*}

\section{Introduction}
\label{sec:intro}
Novel view synthesis (NVS) is a key challenge in computer vision, focusing on rendering images from previously unobserved viewpoints. Recent advancements, such as Neural Radiance Fields (NeRF)~\cite{mildenhall2021nerf} and 3D Gaussian Splatting (3DGS)~\cite{kerbl20233d}, have demonstrated remarkable capabilities in generating novel views using dense captured images of a scene. However, acquiring hundreds to thousands of highly overlapping images of a scene is often time-consuming and impractical, especially for large-scale scenes that require reacquisition whenever changes occur~\cite{Chen2024MultiView3R}. Consequently, there is growing interest in developing efficient approaches to reconstruct 3D scenes from sparse views~\cite{fan2024instantsplat, jiang2024construct, wynn2023diffusionerf, xiong2023sparsegs, wu2024reconfusion, yang2024gaussianobject, paul2024sp2360}.

In the scenario of sparse views, NeRF and 3DGS often overfit due to limited input views, resulting in severe visual artifacts and a lack of coherent structures. 
To tackle this challenge, some existing methods proposed integrating diffusion-based approaches~\cite{rombach2022high} into NVS methods because of their generative capabilities.
These proposed approaches can be categorized into three main types. 
The first type involves using the knowledge of a pre-trained diffusion model and introduce a score distillation sampling (SDS) loss to guide training in the NVS pipeline, e.g.,~\cite{poole2023dreamfusion, xiong2023sparsegs}. The performance of this type of methods, however, are still far from satisfactory in extremely sparse-view cases. 
The second type focuses on generating synthetic images of novel views by training 2D diffusion models on large-scale multi-view datasets. Those synthetic images and reference images are used for training 3D models jointly.
For example, Wu~\textit{et al.}~\cite{wu2024reconfusion} utilized the Latent Diffusion Model (LDM)~\cite{rombach2022high} trained on multiple large-scale datasets to generate images of unobserved views, serving as additional training data during the NeRF model training of scenes. Such methods are time-consuming and costly since they need to be trained on large-scale datasets from scratch. Additionally, the generative training images often contain content not originally present in the scenes due to the nature of generative models, resulting in unexpected elements in some synthesized novel views.
The third type involves fine-tuning diffusion-based models to enhance the visual quality of NVS-rendered images, which are used as additional data for NVS model training. 
Paul~\textit{et al.}~\cite{paul2024sp2360} created a dataset with 3DGS-rendered images and corresponding reference images to fine-tune the Instruct-Pix2Pix diffusion model~\cite{brooks2023instructpix2pix}. During 3DGS training, the rendered images in the current iteration are enhanced by the fine-tuned model and used as training data for the next iteration. 
Yang~\textit{et al.}~\cite{yang2024gaussianobject} adopted a diffusion-based Gaussian repair model to self-generate target images for unobserved views as additional training data for reconstructing extremely sparse-view $360^\circ$ object.
Such approaches rely on accurate ground-truth camera poses, which are often difficult to estimate using typical structure-from-motion methods in extremely sparse-view scenarios.


A few non-diffusion-based NVS methods are proposed to address the challenge of estimating camera poses in sparse-view scenarios.
Fan~\textit{et al.}~\cite{fan2024instantsplat} proposed utilizing DUSt3R~\cite{wang2024dust3r} to estimate camera poses for input images and integrate dense stereo priors with co-visibility relationships, enabling pixel-aligned, progressive expansion of scene geometry without redundancy. Jiang~\textit{et al.}~\cite{jiang2024construct} proposed a method for progressively constructing scenes by utilizing monocular depth to project pixels back into the 3D space, with camera registration and pose adjustments optimized within the pipeline.

In this paper, we present a novel framework for reconstructing 360$^\circ$ scenes in extremely sparse views (four inputs only). We leverage 3DGS for its efficiency in training and rendering as our primary 3D representation method.
To overcome the challenge of estimating camera poses in extremely sparse views, we utilize DUSt3R~\cite{wang2024dust3r} to estimate camera poses and generate a dense point cloud for a scene. We expand more view coverage in 3D space by densely sampling views in the upper hemisphere, where additional training images are rendered using these sampled camera poses and the generated point cloud. Unlike existing methods~\cite{yang2024gaussianobject, paul2024sp2360}, where the camera poses are sampled along the trajectory of the input cameras, our approach provides more comprehensive information of appearance and geometry of scenes from 3D space to train a 3DGS model.
Subsequently, we remove artifacts in the point-cloud-rendered images using a diffusion-based image enhancement model. Retraining the model on our created dataset, our method leverages the generative capabilities of diffusion models to enhance the quality of the point-cloud-rendered images without introducing new content into the images.
The 3DGS model is trained on a combination of the reference images from sparse views and densely sampled synthetic images, enabling the use of more scene information of appearance and geometry for training, yielding the improvement of performance for extremely sparse-view cases.

\section{Method}
\label{sec:method}

\subsection{Preliminary}
\label{ssec:3dgs}

3DGS~\cite{kerbl20233d} is an explicit representation for modeling scenes with parameterized 3D Gaussians. A 3D Gaussian is parameterized by $G=\{\mu, q, s, o, c, sh\}$, where $\mu$ is spatial mean, $q$ is rotation quaternion, $s$ is scaling vector, $o$ is opacity, $c$ is a view-dependent color, and $sh$ is spherical harmonic (SH) coefficients. A 3D scene is represented by a collection of 3D Gaussians $\mathcal{G}=\{G_{t}\}_{t=1}^{K}$, where $K$ is the total number of Gaussians. 
3DGS is a promising technique in the NVS domain due to its efficent integration of structural priors and fast rendering capabilities. Hence, we select 3DGS as the 3D representation method for our study.
To obtain a satisfactory NVS performance, training a 3DGS model typically requires dense view inputs $W\approx 200$~\cite{kerbl20233d,barron2022mip}. In spare view scenario, the number of views $M$ is much less than $W$, resulting in overfitting. To this end, we propose our method to address this challenge.

\begin{table*}[htp]
    \setlength{\tabcolsep}{4.6pt}
    \scriptsize
    \centering
    \begin{tabular}{l|cccc|cccc|cccc}
        \hline
        \multirow{2}{*}{Scene} & \multicolumn{4}{c|}{SSIM $\uparrow$} & \multicolumn{4}{c|}{PSNR $\uparrow$} & \multicolumn{4}{c}{LPIPS $\downarrow$} \\ \cline{2-13} 
        & {\footnotesize Ours} & {\footnotesize InstantSplat} & {\footnotesize COGS} & {\footnotesize DiffusioNeRF} & {\footnotesize Ours} & {\footnotesize InstantSplat} & {\footnotesize COGS} & {\footnotesize DiffusioNeRF} & {\footnotesize Ours} & {\footnotesize InstantSplat} & {\footnotesize COGS} & {\footnotesize DiffusioNeRF} \\ \hline
        Kitchen   & \textbf{0.35} & 0.27 & 0.22 &0.07  & \textbf{14.98} & 14.22 & 11.11 &12.42 & \textbf{0.49} & 0.50 & 0.65 &0.75 \\
        Garden    & \textbf{0.25} & 0.17 & 0.09 &0.06  & \textbf{13.95} & 12.97 & 9.84  &9.03  & 0.61 & \textbf{0.60} & 0.66 &0.94	 \\
        Bonsai    & \textbf{0.38} & 0.27 & 0.24 &0.19  & \textbf{13.64} & 12.53 & 11.18 &12.25 & \textbf{0.56} & 0.58 & 0.63 &0.81 \\
        Bicycle   & \textbf{0.21} & 0.14 & 0.10 &0.08  & \textbf{14.43} & 12.91 & 11.25 &10.73 & 0.64 & \textbf{0.63} & 0.66 &0.87 \\
        Stump     & \textbf{0.27} & 0.17 & 0.15 &0.23  & \textbf{16.15} & 14.22 & 13.88 &12.17 & \textbf{0.63} & \textbf{0.63} &0.62 &0.79 \\
        Treehill  & \textbf{0.30} & 0.22 & 0.18 &0.15  & \textbf{14.53} & 13.78 & 11.47 &12.29 & \textbf{0.59} & \textbf{0.59} & 0.62 &0.85	 \\
        Family    & \textbf{0.43} & 0.38 & 0.22 &0.03  & \textbf{12.20} & 11.80 & 9.28  &9.34 & \textbf{0.50} & 0.51 & 0.63 &0.81	 \\
        Horse     & \textbf{0.55} & 0.52 & 0.33 &0.08 & \textbf{12.83}  & 12.53 & 8.24  &8.96 & \textbf{0.46} & \textbf{0.46} & 0.63 &0.79 \\
        Francis   & \textbf{0.57} & 0.49 & 0.34 &0.09  & \textbf{13.44} & 12.41 & 10.07 &11.60 & \textbf{0.49} & 0.50 & 0.57 &0.74 \\ \hline
        Avg.      & \textbf{0.37} & 0.29 & 0.21 &0.11 & \textbf{14.02} & 13.04 & 10.70 &10.98 & \textbf{0.55} & 0.56 & 0.63 &0.82  \\ \hline
    \end{tabular}
    \caption{Quantitative comparison of our proposed method with benchmarking sparse-view NVS methods. The best performance values for each metric are highlighted, a convention that is continued in all subsequent sections.}
    \label{tab:result_metrics}
\end{table*}

\subsection{Overview of the proposed framework}
\label{ssec:framework}
Given a sparse set of $M$ reference images $X_{\text{r}}\!=\!\{x_{\text{r}}^i\}_{i=1}^M$ of a scene, captured over a $360^\circ$ range, our objective is to obtain a 3D representation $\mathcal{G}$ to achieve photorealistic rendering $x=\mathcal{G}(\pi\!\mid\!\{x_{\text{r}}^i\}_{i=1}^M)$ from any viewpoint $\pi$. 
The proposed framework is illustrated in \mbox{Fig.~\ref{fig:framework}}. 
Initially, the off-the-shelf
multi-view stereo method DUSt3R~\cite{wang2024dust3r} is used to estimate the camera poses $\Pi_{\text{r}}\!=\!\{\pi_{\text{r}}^i\}_{i=1}^M$ of the reference images and generate a dense point cloud.
Subsequently, we sample $N$ camera poses $\Pi_{\text{s}} = \{\pi_{\text{s}}^j\}_{j=1}^N$ in the upper hemisphere space of the scene based on $\Pi_{\text{r}}$. Then, we render images with $\Pi_{\text{s}}$ and the dense point cloud, yielding a point-cloud-rendered image set $X_{\text{p}}\!=\!\{x_{\text{p}}^j\}_{j=1}^N$. Next, the quality of $X_{\text{p}}$ are enhanced by a diffusion-based model retrained on our created dataset, yielding a synthetic image set $X_{\text{s}}\!=\!\{x_{\text{s}}^j\}_{j=1}^N$. Finally, the 3DGS model is trained on $X_{\text{r}}$ and $X_{\text{s}}$.


\subsection{Sampling views in the upper hemisphere space}
After applying the DUSt3R to estimate camera poses of a scene, we sample views within the upper hemisphere. Specifically, the upper hemisphere space is divided into $L$ elevation levels to enhance both angular and spatial diversity.
Starting with the camera poses of the reference images, we calculate the center and radius $r$ of the hemisphere based on the locations of the cameras. 
We strategically place a decreasing number of views at higher elevations within the levels, employing a ratio $\tau$ to constrain the maximum elevation angle. 
Additionally, to align with the setup of the input views, all sampled cameras are oriented to point directly toward the center of the scene.

At each elevation level $l$, the number of cameras $k$ is determined by a uniform distribution derived from the Fibonacci sequence $F(T_l)$ and a predefined set $Q = \{3,4, \ldots, T_l\}_{l=1}^L$. A camera position in spherical coordinates is then parameterized as $\pi_{\text{s}} = \{r, \theta_{(k,l)}, \varphi_{(k,l)}\}$.
Here $\theta_{(k, l)}$ and $\varphi_{(k, l)}$ are calculated as,
\begin{align}
    \theta_{(k, l)} &= \frac{2\pi (k-1)}{F(T_l)}, & \varphi_{(k, l)} &= \frac{\tau \pi}{2L} \times (L - l + 1),
\end{align}
where $l$ ranges from 1 to $L$, and $k$ ranges from 1 to $F(T_l)$. 
Given $N$ sampled camera poses $\Pi_{\text{s}} = \{\pi_{\text{s}}^j\}_{j=1}^N$ and the point cloud, we obtain a rendered image set $X_{\text{p}}$. As DUSt3R-created point cloud is created using corresponding point maps of input images, the overlapping regions of the point maps often contain specific artifacts.

\subsection{Enhancing point-cloud-rendered images}\label{ssec:enhance}
To remove the point-cloud rendering artifacts from the images, we adopt a specialized image restoration framework DiffBIR~\cite{lin2023diffbir}. It is a two-stage architecture designed to address complex rendering artifacts without introducing undesired details. 
Specifically, we retrain its SwinIR~\cite{liang2021swinir} module to address specific artifacts in point cloud rendered images. 
We create our training dataset from three large-scale multi-view datasets, i.e., WildRGB-D~\cite{xia2024rgbd}, MVImgNet~\cite{yu2023mvimgnet}, and DL3DV-10K~\cite{ling2024dl3dv}. For a scene in the datasets, we sample sparse views to create a point cloud using DUSt3R~\cite{wang2024dust3r}. 
The point cloud is then used to render synthetic images for the remaining views. 
Those synthetic images
paired with corresponding reference images serves as the training data.
As a result, 52,552 image pairs are collected as training data.
Due to imprecise camera poses estimated by DUSt3R, which often result in misalignment between rendered and ground-truth images, we incorporate contextual loss from~\cite{mechrez2018contextual} along with the default MSE loss in DiffBIR to guide training. 
The retrained model apply to point-cloud-rendered image set $X_{\text{p}}$ and generate the enhanced synthetic image set $X_{\text{s}}$.


\begin{figure*}[ht]
    \centering
    \includegraphics[width=0.95\textwidth]{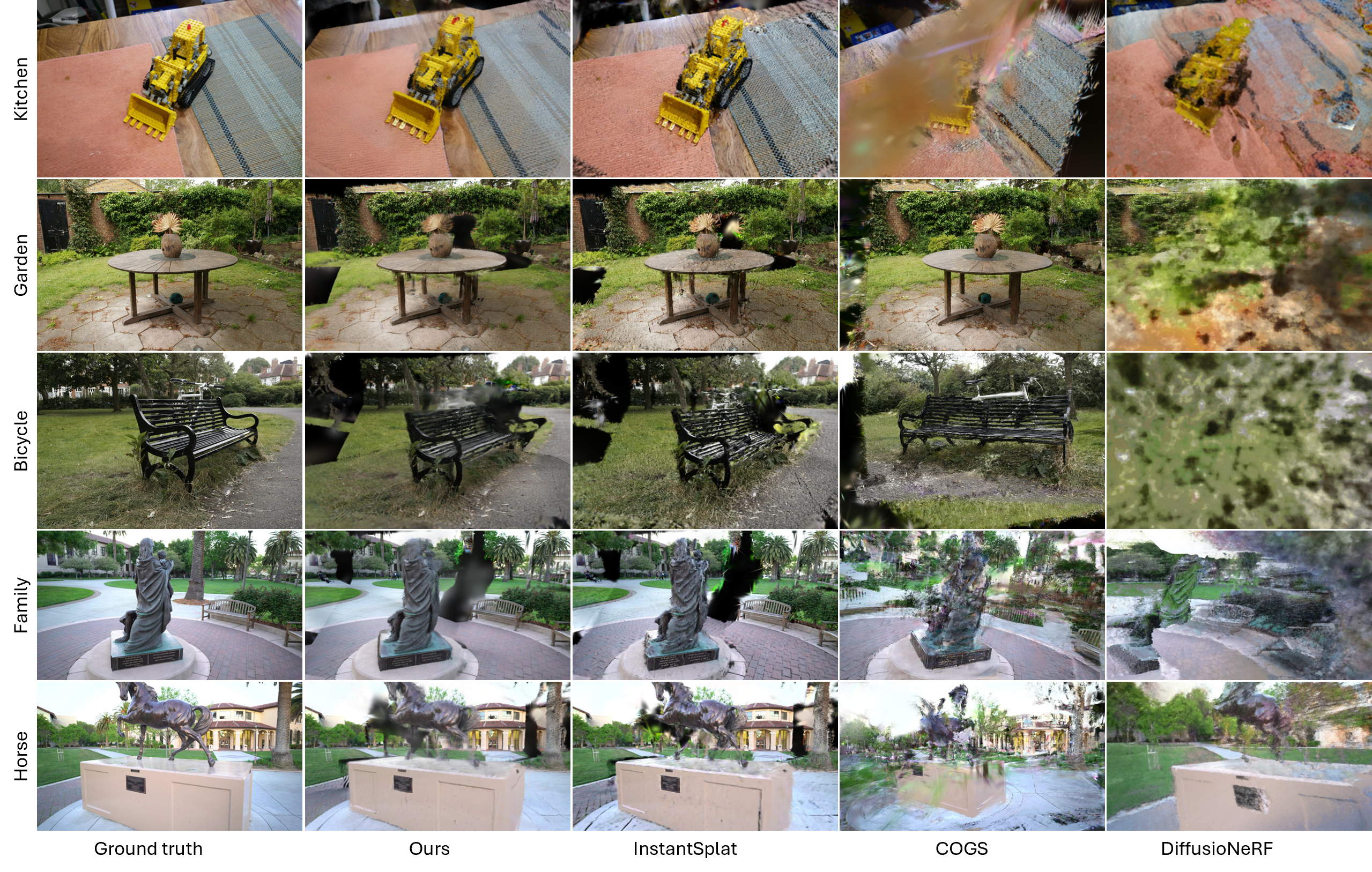}
  \caption{Qualitative comparison of our proposed method with benchmarking sparse-view NVS methods. Each row represents a scene, and each column represents a method or the ground truth.
  }
  \label{fig:result_figure}
\vspace{-10pt}
\end{figure*}

\subsection{Training 3DGS with reference and synthetic images}
\label{ssec:opt}
The 3DGS model is trained jointly on $X_{\text{s}}$ and $X_{\text{r}}$, where its  overall loss~$\mathcal{L}_{\text{all}}$ combines the RGB loss for reference images $X_{\text{r}}$ and synthetic images $X_{\text{s}}$, respectively. 
Let $\hat{x}$ be the corresponding 3DGS-rendered images, $\mathcal{L}_{\text{all}}$ is defined as:

\begin{equation}
    \begin{aligned}
        \mathcal{L}_{\text{all}} = \mathcal{L}_{\text{rgb}}&(x^{i}_{\text{r}}, \hat{x}_{\text{r}}^{i}) +\lambda(\pi_{\text{s}}^j) \mathcal{L}_{\text{rgb}}(x^{j}_{\text{s}}, \hat{x}_{\text{s}}^{j}), \\
        &\quad i \in \{{1,..., M\}, \quad j \in \{1,..., N\}}.
    \end{aligned}
    \label{eqn:loss_synth}
\end{equation}

Here, $\lambda(\pi_{\text{s}}^j)$ serves as a distance-aware weight~\cite{yang2024gaussianobject}, emphasizing that synthetic views closer to reference images contain more scene-relevant appearance information, thus contributing more significantly to the overall loss. This weight is calculated as:
\begin{equation}
    \lambda(\pi_{\text{s}}^j) = 2 \cdot \min_{i=1}^{M} \left( \|\pi_{\text{s}}^j - \pi_{\text{r}}^i\|_2 \right) / D_{\text{max}},
    \label{eqn:lambda}
\end{equation}
where $D_{\text{max}}$ denotes the maximum distance among all reference camera poses.

The RGB loss $\mathcal{L}_{\text{rgb}}$ combines L1, D-SSIM~\cite{wang2004imagessim}, and perceptual losses~\cite{simonyan2014very}, defined as:
\begin{equation}
    \mathcal{L}_{\text{rgb}}(x, \hat{x}) = 
    (1-\lambda_{\text{S}}) \mathcal{L}_1 
    + \lambda_{\text{S}} \mathcal{L}_{\text{D-SSIM}} 
    + \lambda_{\text{P}} \mathcal{L}_{\text{P}}
    \label{eqn:loss_ref},
\end{equation}
where $\mathcal{L}_{1}=\|x, \hat{x}\|_1$, $\mathcal{L}_{\text{D-SSIM}}=1-\text{SSIM}(x, \hat{x})$, and $\mathcal{L}_{\text{P}}=\text{LPIPS}(x, \hat{x})$. $\lambda_{\text{S}}$ and $\lambda_{\text{P}}$ control the weight of each component.

\section{Experiments}
\label{sec:exp}

\subsection{Dataset}
The two real-world benchmarking datasets featuring large-scale scenes: Mip-NeRF 360~\cite{barron2022mip}, and Tanks\&Temples ($360^\circ$ scenes)~\cite{knapitsch2017tanks} are used to evaluate our method. 
Specifically, we evaluate our method in four-view cases, i.e., $M=4$. To reconstruct a 360$^\circ$ scene, it must have overlap between sampled images, even in extremely spare views. Therefore, we select scenes with a centered object only. As a result, six scenes are selected from Mip-NeRF 360 (kitchen, garden, bonsai, stump, treehill, and bicycle), and three scenes are selected from Tanks\&Temples (family, francis, and horse). 
For each selected scene, we adopt the original 3DGS~\cite{kerbl20233d} splitting strategy to partition the data into training and test sets. For the training set, we select four views that ensure coverage of a 360$^\circ$ range around the center of the scene and maintain both angular and spatial diversity in camera poses. We keep the test set for evaluation. 
The training images, and the camera distribution are illustrated in the supplementary materials\footnote{\url{https://sigport.org/sites/default/files/docs/ICIP2025_Sparse_3D_Suppl_Final.pdf}}.




\subsection{Metrics}
Three widely used metrics are applied to evaluate the performance of NVS, including the Structural Similarity Index Measure (SSIM)~\cite{wang2004imagessim}, Peak Signal-to-Noise Ratio (PSNR), and Learned Perceptual Image Patch Similarity (LPIPS)~\cite{zhang2018unreasonable}.
Notably, higher values of PSNR and SSIM indicate better performance, whereas a lower LPIPS score is preferable.

\subsection{Implementation details}
Our framework, shown in Fig.~\ref{fig:framework}, builds on the 3DGS~\cite{kerbl20233d} and InstantSplat~\cite{fan2024instantsplat} codebases. We trained the 3DGS model for 1,000 iterations with $\lambda_{\text{S}}$ set to 0.2, following~\cite{kerbl20233d}. Perceptual loss weights $\lambda_{\text{P}}$ were set to 0.5 for reference images ($X_{\text{r}}$) and 0.1 for synthetic images ($X_{\text{s}}$).
To generate synthetic images, we set $\tau$ to 0.8 imprically, and segmented the upper hemisphere space into five elevation levels, i.e., $L=5$. As a result, 50 additional synthetic training images were created for each scene. Image enhancement was conducted using the retrained \mbox{DiffBIR} model~\cite{lin2023diffbir} with 5 DDIM~\cite{rombach2022high} sampling steps. For evaluation, the resolution scale was set to 4 for the Mip-NeRF 360 dataset and 2 for the Tanks\&Temples dataset.
Since the ground-truth camera poses of the test images are derived from COLMAP using the entire dataset, while our pipeline generates a point cloud using DUSt3R from only four images, resulting in a different camera pose coordinate system~\cite{fan2024instantsplat}. Thus, we used iComma~\cite{sun2023icomma} to register the camera poses of the test images.

\subsection{Baselines}
We compare the performance of our proposed method with two state-of-the-art pose-free methods tailored for sparse-view scenarios, namely, InstantSplat~\cite{fan2024instantsplat} and COGS~\cite{jiang2024construct}. Additionally, we compare our approach with DiffusioNeRF~\cite{wynn2023diffusionerf}, a diffusion-based NVS technique optimized for sparse views that utilizes ground-truth camera poses obtained from COLMAP~\cite{schonberger2016structure} as input.

\subsection{Result}
\label{sec:result}
The results of quantitative comparison presented in Table~\ref{tab:result_metrics}, indicating that our approach outperforms benchmarking methods across all performance metrics. Specifically, our method achieves improvements of 0.98 in PSNR, 0.08 in SSIM, and a slight gain of 0.01 in LPIPS compared to the baseline method, InstantSplat.

Fig.~\ref{fig:result_figure} presents the results of qualitative comparison.
It shows that DiffusioNeRF struggles to reconstruct 360$^\circ$ scenes in our extremely sparse-view setup due to difficulties in establishing correspondences between widely spaced images, causing geometric ambiguities. COGS, which autonomously estimates camera poses without predefined inputs, fails to accurately estimate camera positions in 360$^\circ$ scenes, resulting in test images rendered from incorrect viewpoints. In contrast, InstantSplat and our method leverage DUSt3R to generate camera poses and point clouds as inputs, enabling effective handling of 360$^\circ$ scenes, even in an extremely sparse-view setup. Our method further introduces more scene information of apperance and geometry by using upper hemisphere sampled synthetic images as additional training data, thus achieving better performance than InstantSplat.
In addition, due to the extremely sparse nature of our setup, large background areas remain unrepresented in the point cloud, leading to some empty regions appearing in our results.
Despite these empty regions, our method still outperforms the benchmarking approaches.

\subsection{Ablation study} \label{sec:abl} 
We conducted ablation studies to assess the contribution of each component of our method. Here, the average performance metrics across all scenes are reported.

\textbf{Impact of method components:} Using InstantSplat as the baseline, which relies solely on reference images for training, the results in Table~{\ref{tab:abl_exp_full}} demonstrate a significant PSNR improvement of 0.54 when synthetic images are used as additional training data. Incorporating perceptual loss $\mathcal{L}_{\text{P}}$ , and distance-weighting $\lambda(\pi)$  further improves PSNR and SSIM by 0.19 and 0.02, respectively. Moreover, adding an image enhancement step yields a notable PSNR increase of 0.25. These results indicate the effectiveness of each components.

\begin{table}[ht]
\setlength{\tabcolsep}{5pt}
\scriptsize
\centering
\caption{Ablation study on the components in our method. `Enh.' denotes the enhancement of synthetic images.}
\begin{tabular}{ccccc|c|c|c}
\hline
Reference &Synthetic & $\mathcal{L}_{\text{P}}$  & $\lambda(\pi)$  & Enh.  & SSIM $\uparrow$ & PSNR $\uparrow$  & LPIPS $\downarrow$ \\ \hline
\checkmark            &               &             &     &            &0.29 &13.04 &0.56   \\
\checkmark  &\checkmark     &             &               &            &0.33 &13.58 &0.56   \\
\checkmark  &\checkmark     &\checkmark   &\checkmark     &            &0.35 &13.77 &\textbf{0.55}   \\
\checkmark  &\checkmark     &\checkmark   &\checkmark     &\checkmark  &\textbf{0.37} &\textbf{14.02} &\textbf{0.55}   \\ \hline
\end{tabular}
\label{tab:abl_exp_full}
\end{table}

\textbf{Impact of sampling views in upper hemisphere:} We compare our method with one that samples views along the camera trajectory. Each method samples 50 views as in the main experiment. The results in Table~\ref{tab:abl_exp_hemisphere} show an improvement in PSNR by 0.18 and SSIM by 0.01 when using our method, which indicates the effectiveness of sampling views in the upper hemisphere.

\begin{table}[ht]
\setlength{\tabcolsep}{15pt}
\scriptsize
\centering
\caption{Ablation study on synthetic views sampling strategies.}
\begin{tabular}{l|c|c|c}
\hline
Sampling strategy   & SSIM $\uparrow$ & PSNR $\uparrow$  & LPIPS $\downarrow$ \\ \hline
Camera trajectory   & 0.36            & 13.84            & \textbf{0.55}  \\
Upper hemisphere    & \textbf{0.37}   & \textbf{14.02}   & \textbf{0.55}  \\ \hline
\end{tabular}
\label{tab:abl_exp_hemisphere}
\end{table}

\textbf{Impact of contextual loss for retraining image enhancement module:} 
We compare two loss functions when retraining the SwinIR model within the DiffBIR framework on our created dataset: the default MSE loss alone, and a combination of MSE and contextual loss.  
The results, reported in Table~\ref{tab:abl_exp_diffbir}, indicate that adding contextual loss significantly improves all metrics, increasing PSNR by 0.53, SSIM by 0.02, and reducing LPIPS by 0.01. It demonstrates the efficacy of contextual loss in handling misalignment between point-cloud-rendered and corresponding reference images during the training on our created dataset.

\begin{table}[ht]
\setlength{\tabcolsep}{14.6pt}
\scriptsize
\centering
\caption{Ablation study on different loss functions for retraining the SwinIR model within DiffBIR framework.}
\begin{tabular}{l|c|c|c}
\hline
Loss function     & SSIM $\uparrow$ & PSNR $\uparrow$  & LPIPS $\downarrow$ \\ \hline
Pretrained        & 0.33            & 13.29            & 0.57               \\
MSE Only          & 0.35            & 13.49            & 0.56               \\
Contextual + MSE  & \textbf{0.37}            & \textbf{14.02}            & \textbf{0.55}               \\ \hline
\end{tabular}
\label{tab:abl_exp_diffbir}
\vspace{-10pt}
\end{table}

\section{Conclusion and future work}
\label{sec:future}
We propose a framework to improve the performance of 3DGS for 360$^\circ$ scenes with extremely sparse views. 
The framework uses reference and synthetic images to train a 3DGS model.
Those synthetic images are created with camera poses sampled in the upper hemisphere space of scenes and DUSt3R-created point clouds. 
Moreover, we remove artifacts in synthetic images using a diffusion-based image enhancement model retrained on our created dataset. Experimental results show that our method outperforms benchmarking methods.

In addition to the development of the proposed framework, future work will focus on improving the completeness and quality of the input point cloud. Moreover, improving the NVS performance with the geometry of the point cloud should also be considered. Furthermore, exploring diffusion-based inpainting models to enhance the quality of images rendered from point clouds is also a promising direction. 

\section{Acknowledgment}
This research received partial funding from the Flemish Government under the `Onderzoeksprogramma Artificiele Intelligentie (AI) Vlaanderen' programme.

\vfill\pagebreak

\bibliography{strings,refs}

\begin{thebibliography}{10}

\bibitem{barron2022mip}
J.~T. Barron, B.~Mildenhall, D.~Verbin, P.~P. Srinivasan, and P.~Hedman.
\newblock {Mip-NeRF 360: Unbounded Anti-Aliased Neural Radiance Fields}.
\newblock In {\em IEEE Conference on Computer Vision and Pattern Recognition}, pages 5470--5479, 2022.

\bibitem{brooks2023instructpix2pix}
T.~Brooks, A.~Holynski, and A.~A. Efros.
\newblock {InstructPix2Pix: Learning To Follow Image Editing Instructions}.
\newblock In {\em IEEE Conference on Computer Vision and Pattern Recognition}, pages 18392--18402, 2023.

\bibitem{Chen2024MultiView3R}
G.~Chen, M.~Vlaminck, W.~Philips, and H.~Q. Luong.
\newblock {Multi-View 3D Reconstruction for Construction Site Monitoring}.
\newblock In {\em Conference on VISAPP}, 2024.

\bibitem{fan2024instantsplat}
Z.~Fan, W.~Cong, K.~Wen, K.~Wang, J.~Zhang, X.~Ding, D.~Xu, B.~Ivanovic, M.~Pavone, G.~Pavlakos, et~al.
\newblock {InstantSplat: Unbounded Sparse-view Pose-free Gaussian Splatting in 40 Seconds}.
\newblock {\em arXiv preprint arXiv:2403.20309}, 2024.

\bibitem{jiang2024construct}
K.~Jiang, Y.~Fu, M.~Varma~T, Y.~Belhe, X.~Wang, H.~Su, and R.~Ramamoorthi.
\newblock {A Construct-Optimize Approach to Sparse View Synthesis without Camera Pose}.
\newblock In {\em ACM SIGGRAPH 2024 Conference Papers}, pages 1--11, 2024.

\bibitem{kerbl20233d}
B.~Kerbl, G.~Kopanas, T.~Leimk{\"u}hler, and G.~Drettakis.
\newblock {3D Gaussian Splatting for Real-Time Radiance Field Rendering}.
\newblock {\em ACM Transactions on Graphics}, 42(4):139--1, 2023.

\bibitem{knapitsch2017tanks}
A.~Knapitsch, J.~Park, Q.-Y. Zhou, and V.~Koltun.
\newblock {Tanks and temples: benchmarking large-scale scene reconstruction}.
\newblock {\em ACM Transactions on Graphics}, 36(4):1--13, 2017.

\bibitem{liang2021swinir}
J.~Liang, J.~Cao, G.~Sun, K.~Zhang, L.~Van~Gool, and R.~Timofte.
\newblock {SwinIR: Image Restoration Using Swin Transformer}.
\newblock In {\em IEEE International Conference on Computer Vision}, pages 1833--1844, 2021.

\bibitem{lin2023diffbir}
X.~Lin, J.~He, Z.~Chen, Z.~Lyu, B.~Dai, F.~Yu, W.~Ouyang, Y.~Qiao, and C.~Dong.
\newblock {DiffBIR: Toward Blind Image Restoration with Generative Diffusion Prior}.
\newblock {\em arXiv preprint arXiv:2308.15070}, 2023.

\bibitem{ling2024dl3dv}
L.~Ling, Y.~Sheng, Z.~Tu, W.~Zhao, C.~Xin, K.~Wan, L.~Yu, Q.~Guo, Z.~Yu, Y.~Lu, et~al.
\newblock {DL3DV-10K: A Large-Scale Scene Dataset for Deep Learning-based 3D Vision}.
\newblock In {\em IEEE Conference on Computer Vision and Pattern Recognition}, pages 22160--22169, 2024.

\bibitem{mechrez2018contextual}
R.~Mechrez, I.~Talmi, and L.~Zelnik-Manor.
\newblock {The Contextual Loss for Image Transformation with Non-Aligned Data}.
\newblock In {\em IEEE Conference on European Conference on Computer Vision}, pages 768--783, 2018.

\bibitem{mildenhall2021nerf}
B.~Mildenhall, P.~P. Srinivasan, M.~Tancik, J.~T. Barron, R.~Ramamoorthi, and R.~Ng.
\newblock Nerf: Representing scenes as neural radiance fields for view synthesis.
\newblock {\em Communications of the ACM}, 65(1):99--106, 2021.

\bibitem{paul2024sp2360}
S.~Paul, C.~Wewer, B.~Schiele, and J.~E. Lenssen.
\newblock {$Sp^2360$: Sparse-view $360^\circ$ Scene Reconstruction using Cascaded 2D Diffusion Priors}.
\newblock In {\em ECCV 2024 Workshop on Wild 3D: 3D Modeling, Reconstruction, and Generation in the Wild}, 2024.

\bibitem{poole2023dreamfusion}
B.~Poole, A.~Jain, J.~T. Barron, and B.~Mildenhall.
\newblock {DreamFusion: Text-to-3D using 2D Diffusion}.
\newblock In {\em International Conference on Learning Representations}, 2023.

\bibitem{rombach2022high}
R.~Rombach, A.~Blattmann, D.~Lorenz, P.~Esser, and B.~Ommer.
\newblock {High-Resolution Image Synthesis With Latent Diffusion Models}.
\newblock In {\em IEEE Conference on Computer Vision and Pattern Recognition}, pages 10684--10695, 2022.

\bibitem{schonberger2016structure}
J.~L. Schonberger and J.-M. Frahm.
\newblock {Structure-From-Motion Revisited}.
\newblock In {\em IEEE Conference on Computer Vision and Pattern Recognition}, pages 4104--4113, 2016.

\bibitem{simonyan2014very}
K.~Simonyan.
\newblock Very deep convolutional networks for large-scale image recognition.
\newblock {\em arXiv preprint arXiv:1409.1556}, 2014.

\bibitem{sun2023icomma}
Y.~Sun, X.~Wang, Y.~Zhang, J.~Zhang, C.~Jiang, Y.~Guo, and F.~Wang.
\newblock {iComMa: Inverting 3D Gaussian Splatting for Camera Pose Estimation via Comparing and Matching}.
\newblock {\em arXiv preprint arXiv:2312.09031}, 2023.

\bibitem{wang2024dust3r}
S.~Wang, V.~Leroy, Y.~Cabon, B.~Chidlovskii, and J.~Revaud.
\newblock {DUSt3R: Geometric 3D Vision Made Easy}.
\newblock In {\em IEEE Conference on Computer Vision and Pattern Recognition}, pages 20697--20709, 2024.

\bibitem{wang2004imagessim}
Z.~Wang, A.~C. Bovik, H.~R. Sheikh, and E.~P. Simoncelli.
\newblock {Image quality assessment: from error visibility to structural similarity}.
\newblock {\em IEEE Transactions on Image Processing}, 13(4):600--612, 2004.

\bibitem{wu2024reconfusion}
R.~Wu, B.~Mildenhall, P.~Henzler, K.~Park, R.~Gao, D.~Watson, P.~P. Srinivasan, D.~Verbin, J.~T. Barron, B.~Poole, et~al.
\newblock {ReconFusion: 3D Reconstruction with Diffusion Priors}.
\newblock In {\em IEEE Conference on Computer Vision and Pattern Recognition}, pages 21551--21561, 2024.

\bibitem{wynn2023diffusionerf}
J.~Wynn and D.~Turmukhambetov.
\newblock {DiffusioNeRF: Regularizing Neural Radiance Fields With Denoising Diffusion Models}.
\newblock In {\em IEEE Conference on Computer Vision and Pattern Recognition}, pages 4180--4189, 2023.

\bibitem{xia2024rgbd}
H.~Xia, Y.~Fu, S.~Liu, and X.~Wang.
\newblock {RGBD Objects in the Wild: Scaling Real-World 3D Object Learning from RGB-D Videos}.
\newblock In {\em IEEE Conference on Computer Vision and Pattern Recognition}, pages 22378--22389, 2024.

\bibitem{xiong2023sparsegs}
H.~Xiong, S.~Muttukuru, R.~Upadhyay, P.~Chari, and A.~Kadambi.
\newblock {SparseGS: Real-Time $360^\circ$ Sparse View Synthesis using Gaussian Splatting}.
\newblock {\em arXiv preprint arXiv:2312.00206}, 2023.

\bibitem{yang2024gaussianobject}
C.~Yang, S.~Li, J.~Fang, R.~Liang, L.~Xie, X.~Zhang, W.~Shen, and Q.~Tian.
\newblock {GaussianObject: High-Quality 3D Object Reconstruction from Four Views with Gaussian Splatting}.
\newblock {\em arXiv preprint arXiv:2402.10259}, 2024.

\bibitem{yu2023mvimgnet}
X.~Yu, M.~Xu, Y.~Zhang, H.~Liu, C.~Ye, Y.~Wu, Z.~Yan, C.~Zhu, Z.~Xiong, T.~Liang, et~al.
\newblock {MVImgNet: A Large-Scale Dataset of Multi-View Images}.
\newblock In {\em IEEE Conference on Computer Vision and Pattern Recognition}, pages 9150--9161, 2023.

\bibitem{zhang2018unreasonable}
R.~Zhang, P.~Isola, A.~A. Efros, E.~Shechtman, and O.~Wang.
\newblock {The Unreasonable Effectiveness of Deep Features as a Perceptual Metric}.
\newblock In {\em IEEE Conference on Computer Vision and Pattern Recognition}, pages 586--595, 2018.

\end{thebibliography}

\end{document}